\documentclass{article}

    \usepackage[preprint]{neurips_2019}

\usepackage[utf8]{inputenc} 
\usepackage[T1]{fontenc}    
\usepackage{hyperref}       
\usepackage{url}            
\usepackage{booktabs}       
\usepackage{amsfonts}       
\usepackage{nicefrac}       
\usepackage{microtype}      

\usepackage[pdftex]{graphicx}
\usepackage{amsmath}        
\usepackage{xcolor}         
\usepackage[skip=2pt]{caption}
\usepackage{enumitem}

\usepackage{graphicx}
\usepackage{amsmath}
\usepackage{amssymb}
\usepackage{booktabs}
\usepackage{multirow}
\usepackage{makecell}
\usepackage{fixltx2e}
\usepackage{tabularx}
\usepackage{array}
\usepackage{caption} 

\newcolumntype{L}{>{\centering\arraybackslash}m{3cm}}
\newcommand{\eg}{\emph{e.g.}}
\newcommand{\ie}{\emph{i.e.}}

\title{Adversarial Distillation for Ordered Top-$k$ Attacks}

\author{
  Zekun Zhang and Tianfu Wu\thanks{Corresponding author.} \\
  Department of Electrical and Computer Engineering and the Visual Narrative Initiative\\
  North Carolina State University\\
  Raleigh, NC 27695 \\
  \texttt{\{zzhang56, tianfu\_wu\}@ncsu.edu} \\
}

\begin{document}

\maketitle



\begin{abstract}
  Deep Neural Networks (DNNs) are vulnerable to adversarial attacks, especially white-box targeted attacks. One scheme of learning attacks is to design a proper adversarial objective function that leads to the imperceptible perturbation for any test image (\eg, the Carlini-Wagner (C\&W) method~\cite{CWAttack}). Most methods address targeted attacks in the Top-$1$ manner. In this paper, we propose to learn \textbf{ordered Top-$k$ attacks} ($k\geq 1$) for image classification tasks, that is to enforce the Top-$k$ predicted labels of an adversarial example to be the $k$ (randomly) selected and ordered labels (the ground-truth label is exclusive). To this end, we present an \textbf{adversarial distillation} framework: First, we compute an adversarial probability distribution for any given ordered Top-$k$ targeted labels with respect to the ground-truth of a test image. Then, we learn  adversarial examples by minimizing the Kullback-Leibler (KL) divergence together with the perturbation energy penalty, similar in spirit to the network distillation method~\cite{distillation}. We explore how to leverage label semantic similarities  in computing the targeted distributions, leading to \textbf{knowledge-oriented attacks}.
  In experiments, we thoroughly test Top-$1$ and Top-$5$ attacks in the ImageNet-1000~\cite{ImageNet} {\tt validation} dataset using two popular DNNs trained with clean ImageNet-1000 {\tt train} dataset, ResNet-50~\cite{ResidualNet} and DenseNet-121~\cite{DenseNet}. For both models, our proposed adversarial distillation approach outperforms the C\&W method in the Top-$1$ setting, as well as other baseline methods. Our approach shows significant improvement in the Top-$5$ setting against a strong modified C\&W method.
\end{abstract}

\section{Introduction}
Visual recognition systems (\eg, image classification) play key roles in a wide range of applications such as autonomous driving, robot autonomy, smart medical diagnosis and video surveillance. Recently, remarkable progress has been made through big data and powerful GPUs driven deep neural networks (DNNs) under supervised learning framework~\cite{LeCunCNN,AlexNet}. DNNs significantly increase prediction accuracy in visual recognition tasks and even outperform humans in some image classification tasks~\cite{ResidualNet,InceptionNet}.
Despite the dramatic improvement, it has been shown that DNNs trained for visual recognition tasks can be easily fooled by so-called \textbf{adversarial attacks} which utilize visually imperceptible, carefully-crafted perturbations to cause networks to misclassify inputs in arbitrarily chosen ways in the close set of labels used in training~\cite{FoolDeepNet,LBFGS, AdversarialExampl, CWAttack}, even with one-pixel attack~\cite{OnePixelAttack}. Assuming full access to DNNs pretrained with clean images, white-box targeted attacks are powerful ways of investigating the brittleness of DNNs and their sensitivity to non-robust yet well-generalizing features in the data, and of exploiting adversarial examples as useful features~\cite{AttackNotBugs}. 

\begin{figure} [t]
  \centering
    \includegraphics[width=1.0\linewidth]{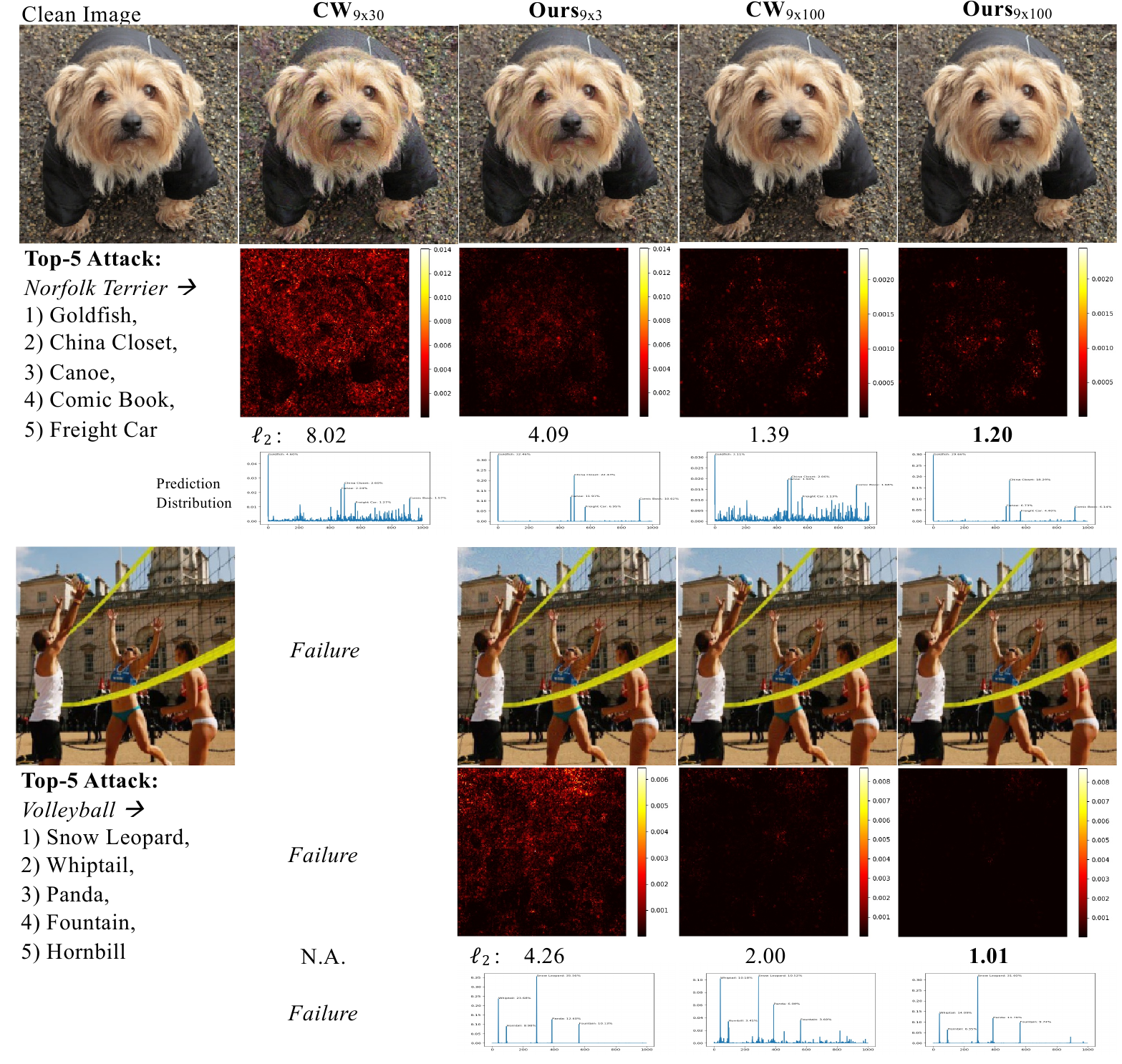}
  \caption{\small Examples of ordered Top-$5$ attacks for a pretrained ResNet-50 model~\cite{ResidualNet} in the ImageNet-1000 dataset. We compare a modified C\&W method~\cite{CWAttack} (see details in Sec.~\ref{sec:modifiedCW}) and our proposed adversarial distillation method in terms of $\ell_2$ distance between a clean image and the learned adversarial example. 
  Here, ${9\times30}$ and ${9\times1000}$ refer to the settings of hyperparameter search: we perform $9$ binary searches for the trade-off parameter of perturbation energy penalty in the objective function and each search step takes $30$ and $1000$ iterations of optimization respectively. Our method consistently outperforms the modified C\&W method. Similarly, our method obtains better results for the bottom image of \textit{Volleyball}, where the modified C\&W$_{9\times30}$ method  fails to attack. (Best viewed in color and magnification)}\label{fig:ex-top5-attack}
\end{figure}

In this paper, we focus on learning  visually-imperceptible targeted attacks under the white-box setting. One scheme of learning these attacks is to design a proper adversarial objective function that leads to the imperceptible perturbation for any test image, \eg, the widely used Carlini-Wagner (C\&W) method~\cite{CWAttack}. However, most methods address targeted attacks in the Top-$1$ manner, which limits the  flexibility of attacks, and may lead to less rich perturbations. We propose to generalize this setting to account for \textbf{ordered Top-$k$ targeted attacks}, that is to enforce the Top-$k$ predicted labels of an adversarial example to be the $k$ (randomly) selected and ordered labels ($k\geq 1$, the ground-truth (GT) label is exclusive). Figure~\ref{fig:ex-top5-attack} shows two examples. 

To see why Top-$k$ targeted attacks are entailed, let's take a close look at the ``robustness" of an attack method itself under the traditional Top-$1$ protocol. One crucial question is, 

\textit{How far is the attack method able to push the underlying ground-truth label in the prediction of the learned adversarial examples?} 

Consider a white-box targeted attack method such as the C\&W method~\cite{CWAttack}. Although it can achieve $100$\% attack success rate (ASR) under the given Top-$1$ protocol, if the ground-truth labels of adversarial examples still largely appear in the Top-$5$ of the prediction, we may be over-confident about the $100$\% ASR, especially when some downstream modules may rely on Top-$5$ predictions in their decision making. Table~\ref{tab:gt-rank} shows the results. The C\&W method does not push the GT labels very far, especially when smaller perturbation energy is aimed using larger search range (\eg, the average rank of the GT label is $2.6$ for C\&W$_{9\times1000}$). On the contrary, the three untargeted attack approaches work much better in terms of pushing the GT labels, although their perturbation energy are usually much larger. 
What is interesting to us is the difference of the objective functions used by the C\&W method and the three untargeted attack methods respectively. The former maximizes the margin of the logits between the target and the runner-up (either GT or not), while the latter maximizes the cross-entropy between the prediction probabilities (softmax of logits) and the one-hot distribution of the ground-truth. Furthermore, the label smoothing methods~\cite{LabelSmoothing, ConfidencePenalty} is often used to improve the performance of DNNs, which address the over-confidence in the one-hot vector encoding of annotations. And, the network distillation method~\cite{distillation,distillation1} views the knowledge of a DNN as the conditional distribution it produces over outputs given an input. One question naturally arises,

\textit{Can we design a proper adversarial distribution similar in spirit to label smoothing to guide the ordered Top-$k$ attack by leveraging the view of point of network distillation?}

Our proposed method aims to harness the best of the above strategies  in designing proper target distributions and objective functions to achieve both high ASR and low perturbation energy. Our proposed ordered Top-$k$ attacks explicitly push the GT labels to a ``safe" zone of retaining the ASR.


\begin{table}
\caption{\small Results of showing where the ground-truth (GT) labels are in the prediction of learned adversarial examples for different attack methods. The test is done in ImageNet-1000 {\tt validation} dataset using a pretrained ResNet-${50}$ model~\cite{ResidualNet}. Please see Sec.~\ref{sec:exp-setup} for details of experimental settings. }
\label{tab:gt-rank}
\centering
\resizebox{0.8\textwidth}{!}{
\begin{tabular}{llllllll} 
\toprule
\multirow{2}{*}{Method} &\multirow{2}{*}{ASR} & \multicolumn{5}{c}{Proportion of GT Labels in Top-$k$ {\small (smaller is better)}}&\multirow{2}{3cm}{{\footnotesize Average Rank of GT Labels  (larger is better)}}\\  \cmidrule(r){3-7}
&&Top-$3$ &Top-${5}$ &Top-${10}$ &Top-${50}$ &Top-${100}$\\ \midrule
C\&W$_{9\times30}$~\cite{CWAttack} &99.9   &36.9  &50.5  &66.3  &90.0  &95.1 &20.4\\
C\&W$_{9\times1000}$~\cite{CWAttack} &100&71.9&87.0&96.1&99.9&100&2.6\\ 
\hline 
FGSM~\cite{FGSM} &80.7 &25.5&37.8  &52.8  &81.2  &89.2  &44.2\\
PGD$_{10}$~\cite{IFGSM, PGD} &100   &3.3  &6.7  &12  &34.7  &43.9 &306.5\\
MIFGSM$_{10}$~\cite{MIFGSM} &99.9  &0.7  &1.9  &6.0  &22.5  &32.3   &404.4\\
\bottomrule
\end{tabular}
}
\end{table}


Towards learning the generalized ordered Top-$k$ attacks, we present an \textbf{adversarial distillation} framework: First, we compute an adversarial probability distribution for any given ordered Top-$k$ targeted labels with respect to the ground-truth of a test image. Then, we learn  adversarial examples by minimizing the Kullback-Leibler (KL) divergence together with the perturbation energy penalty, similar in spirit to the network distillation method~\cite{distillation}. More specifically, we explore how to leverage label semantic similarities  in computing the targeted distributions, leading to \textbf{knowledge-oriented attacks}. We measure  label  semantic similarities using the cosine distance between some off-the-shelf word2vec embedding of labels such as the pretrained Glove embedding~\cite{Glove}.
Along this direction, a few questions of interest that naturally arise are studied: Are all Top-$k$ targets equally challenging for an attack approach? How can we leverage the semantic knowledge between labels to guide an attack approach to learn better adversarial examples and to find the weak spots of different attack approaches?  We found that KL is a stronger alternative than the C\&W loss function, and label semantic knowledge is useful in designing effective adversarial distributions. 

In experiments, we develop a modified C\&W approach for ordered Top-$k$ attacks as a strong baseline. We thoroughly test Top-$1$ and Top-$5$ attacks in the ImageNet-1000~\cite{ImageNet} {\tt validation} dataset using two popular DNNs trained with clean ImageNet-1000 {\tt train} dataset, ResNet-50~\cite{ResidualNet} and DenseNet-121~\cite{DenseNet}. For both models, our proposed adversarial distillation approach outperforms the vanilla C\&W method in the Top-$1$ setting, as well as other baseline methods such as the PGD method~\cite{PGD,IFGSM}. Our approach shows significant improvement in the Top-$5$ setting against the modified C\&W method. We observe that Top-$k$ targets that are distant from the GT label in terms of either label semantic distance or prediction scores of clean images are actually more difficulty to attack. 

\textbf{Our Contributions.} This paper makes three main contributions to the field of learning adversarial attacks: 
    (i) To our knowledge, this is the first work of learning ordered Top-$k$ attacks. This generalized setting is a straightforward extension of the widely used Top-$1$ attack and able to improve the robustness of adversarial attacks themselves. 
   (ii) A conceptually simple yet effective framework, adversarial distillation is proposed to learn ordered Top-$k$ attacks under the white-box settings. It outperforms a strong baseline, the C\&W method~\cite{CWAttack} under both the traditional Top-$1$ and the proposed ordered Top-$5$ in the ImageNet-1000 dataset using two popular DNNs, ResNet-50~\cite{ResidualNet} and DenseNet-121~\cite{DenseNet}. 
   (iii) Knowledge-oriented design of adversarial target distributions are studied whose effectiveness is supported by the experimental results. 

\textbf{Paper organization.} The remainder of this paper is organized as follows. In Section~\ref{sec:formulation}, we overview the white-box targeted attacks and the C\&W method, and then present details of our proposed adversarial distillation framework for the ordered Top-$k$ targeted attack. In Section~\ref{sec:exp-setup}, we present thorough comparisons in ImageNet-1000. In Section~\ref{sec:related}, we briefly review the related work. Finally, we conclude this paper in Section~\ref{sec:conclusion}. 

\section{Problem formulation}\label{sec:formulation}
In this section,  we first briefly introduce, to be self-contained, the white-box attack setting and the widely used C\&W method~\cite{CWAttack} under the Top-$1$ protocol. We then define the ordered Top-$k$ attack setting and develop a modified C\&W method for it ($k> 1$)  as a strong baseline. Finally, we present our proposed adversarial distillation framework.   

\subsection{Background on white-box targeted attack under the Top-$1$ setting}
We focus on classification tasks using DNNs. Denote by $(x, y)$ a pair of a clean input $x\in \mathcal{X}$ and its ground-truth label $y\in \mathcal{Y}$. For example, in the ImageNet-1000 classification task, $x$ represents a RGB image defined in the lattice of $224\times 224$ and we have $\mathcal{X}\triangleq R^{3\times 224\times 224}$. $y$ is the category label and we have $\mathcal{Y}\triangleq \{1, \cdots, 1000\}$. Let $f(\cdot;\Theta)$ be a DNN pretrained on clean training data where $\Theta$ collects all estimated parameters and is fixed in learning adversarial examples. For notation simplicity, we denote by $f(\cdot)$ a pretrained DNN. 
The prediction for an input $x$ from $f(\cdot)$ is usually defined using softmax function by,
\begin{equation}
    P =f(x)=softmax(z(x)),
\end{equation}
where $P\in R^{|\mathcal{Y}|}$ represents the estimated confidence/probability vector ($P_c\geq 0$ and $\sum_c P_c=1$) and $z(x)$ is the logit vector. The predicted label is then inferred by $\hat{y}=\arg\max_{c\in [1,|\mathcal{Y}|]} P_c$. 

In learning targeted attacks under the Top-$1$ protocol, for an input $(x, y)$, given a target label $t\neq y$, we seek to compute some visually-imperceptible perturbation $\delta(x, t, f)$  using the pretrained and fixed DNN $f(\cdot)$ under the white-box setting. \textit{White-box attacks} assume the complete knowledge of the pretrained DNN $f$, including its parameter values, architecture, training method, etc. The perturbed example $x'=x+\delta(x, t, f)$ is called \textbf{an adversarial example} of $x$ if $t=\hat{y}'=\arg\max_c f(x')_c$ and the perturbation $\delta(x, t, f)$ is sufficiently small according to some energy metric. We usually focus on the subset of inputs $(x,y)$'s that are correctly classified by the model, \ie, $y=\hat{y}=\arg\max_c f(x)_c$. Learning $\delta(x, t, f)$ under the Top-$1$ protocol is posed as a constrained optimization problem~\cite{AdversarialExampl, CWAttack}, 
\begin{align}
    \text{minimize}\quad &\mathcal{E}(\delta)=||\delta||_p, \label{eq:formulation}\\
    \nonumber \text{subject to}\quad & t=\arg\max_c f(x+\delta)_c,\\
    \nonumber & x+\delta \in [0, 1]^n,
\end{align}
where $\mathcal{E}(\cdot)$ is defined by a $\ell_p$ norm (\eg, the $\ell_2$ norm) and $n$ the size of the input domain (e.g., the number of pixels).  
To overcome the difficulty (non-linear and non-convex constraints) of directly solving Eqn.~\ref{eq:formulation}, the C\&W method expresses it in a different form by designing some loss functions $L(x')=L(x+\delta)$ such that the first constraint $t=\arg\max_c f(x')_c$ is satisfied if and only if  $L(x')\leq 0$. The best loss function proposed by the C\&W method is defined by the hinge loss of logits between the target label and the runner-up, 
\begin{equation}
    L_{CW}(x') = \max(0, \max_{c\neq t}z(x')_c - z(x')_t). \label{eq:cwloss}
\end{equation}
Then, the learning problem becomes, 
\begin{align}
    \text{minimize}\quad & ||\delta||_p + \lambda\cdot  L(x+\delta), \label{eq:formulation1}\\
    \nonumber \text{subject to}\quad & x+\delta \in [0, 1]^n,
\end{align}
which can be solved via back-propagation with the constraint satisfied via introducing a {\tt tanh} layer. For the trade-off parameter $\lambda$, a binary search will be performed during the learning (\eg, $9\times 1000$).  

\subsection{The proposed ordered Top-$k$ attack setting}
It is straightforward to extend Eqn.~\ref{eq:formulation} for learning ordered Top-$k$ attacks ($k\geq 1$). Denote by $(t_1, \cdots, t_k)$ the ordered Top-$k$ targets ($t_i\neq y$). We have, 
\begin{align}
    \text{minimize}\quad &\mathcal{E}(\delta)=||\delta||_p, \label{eq:formulation-topk}\\
    \nonumber \text{subject to}\quad & t_i=\arg\max_{c\in [1, |\mathcal{Y}|], c\notin \{t_1, t_{i-1}\} } f(x+\delta)_c, \quad i\in \{1,\cdots, k\}, \\
    \nonumber & x+\delta \in [0, 1]^n .
\end{align}

\subsubsection{A modified C\&W method}\label{sec:modifiedCW}
We can modify the loss function (Eqn.~\ref{eq:cwloss}) of the C\&W method accordingly to solve Eqn.~\ref{eq:formulation-topk}. We have, 
\begin{equation}
    L^{(k)}_{CW}(x') = \sum_{i=1}^k \max(0, \max_{j\notin \{t_1,\cdots, t_{i}\}}z(x')_j - z(x')_{t_i}). \label{eq:cwloss-topk}
\end{equation}
So, the vanilla C\&W loss (Eqn.~\ref{eq:cwloss}) is the special case of Eqn.~\ref{eq:cwloss-topk} (\ie, when $k=1$).

\subsubsection{Our proposed knowledge-oriented adversarial distillation framework}
In the C\&W loss functions, only the margin of logits between the targeted labels and the runner-ups is taken into account. In our adversarial distillation framework, we adopt the view of point proposed in the network distillation method~\cite{distillation} that the full confidence/probability distribution summarizes the knowledge of a trained DNN. We hypothesize that we can leverage the network distillation framework to learn the ordered Top-$k$ attacks by designing a proper adversarial probability distribution across the entire set of labels that satisfies the specification of the given ordered Top-$k$ targets. 

Consider the Top-$k$ targets, $(t_1, \cdots, t_k)$, we want to define the adversarial probability distribution, denoted by $P^{adv}$ in which $P^{adv}_{t_i}> P^{adv}_{t_j}$ ($\forall i<j$) and $P^{adv}_{t_i}>P^{adv}_j$ ($\forall j\notin (t_1, \cdots, t_k)$). The space of candidate distributions are huge. We present a simple knowledge-oriented approach to define the adversarial distribution. We first specify the logit distribution and then compute the probability distribution using softmax. Denote by $Z$ the maximum logit (\eg, $Z=10$ in our experiments). We define the adversarial logits for the ordered Top-$k$ targets by,
\begin{equation}
    z^{adv}_{t_i}=Z - (i-1)\times \gamma, \quad i\in [1, \cdots, k],
\end{equation}
where $\gamma$ is an empirically chosen decreasing factor (\eg, $\gamma=0.3$ in our experiments). For the remaining categories $j\notin (t_1, \cdots, t_k)$, we define the adversarial logits by,
\begin{equation}
    z^{adv}_j = \alpha \times \frac{1}{k}\sum_{i=1}^k s(t_i, j) + \epsilon, \label{eq:logit-others}
\end{equation}
where $0\leq \alpha < z^{adv}_{t_k}$ is the maximum logit that can be assigned to any $j$, $s(a, b)$ is the semantic similarity between the label $a$ and label $b$, and $\epsilon$ is a small position for numerical consideration (\eg, $\epsilon=1e$-$5$). We compute $s(a, b)$ using the cosine distance between the Glove~\cite{Glove} embedding vectors of category names and $-1\leq s(a, b) \leq 1$. Here, when $\alpha=0$, we discard the semantic knowledge and treat all the remaining categories equally. Note that our design of $P^{adv}$ is similar in spirit to the label smoothing technique and its variants~\cite{LabelSmoothing,ConfidencePenalty} except that we target attack labels and exploit label semantic knowledge. The design choice is still preliminary, although we observe its effectiveness in experiments. We hope this can encourage more sophisticated work to be explored.  

With the adversarial probability distribution $P^{adv}$ defined above as the target, we use the KL divergence as the loss function in our adversarial distillation framework as done in network distillation~\cite{distillation} and we have, 
\begin{equation}
    L^{(k)}_{adv}(x') = KL(f(x')||P^{adv}),
\end{equation}
and then we follow the same optimization scheme as done in the C\&W method (Eqn.~\ref{eq:formulation1}). 

\section{Experiments}\label{sec:exp-setup}
In this section, we present results of our proposed method tested in ImageNet-1000~\cite{ImageNet} using two pretrained DNNs, ResNet-50~\cite{ResidualNet} and DenseNet-121~\cite{DenseNet} from the PyTorch model zoo~\footnote{https://github.com/pytorch/vision/tree/master/torchvision/models}. We implement our method using the AdverTorch toolkit~\footnote{https://github.com/BorealisAI/advertorch}. Our source code will be released.  

\textbf{Data.} In ImageNet-1000~\cite{ImageNet}, there are $50,000$ images for validation. We obtain the subset of images for which the predictions of both the ResNet-50 and DenseNet-121 are correct. To reduce the computational demand, we further test our method in a randomly sampled subset, as commonly done in the literature. To enlarge the coverage of categories, we first randomly select 500 categories and then randomly chose 2 images per selected categories, resulting in 1000 test images in total. 

\textbf{Settings.} We follow the protocol used in the C\&W method. We only test $\ell_2$ norm as the energy penalty for perturbations in learning. But we evaluate learned adversarial examples in terms of three norms ($\ell_1$, $\ell_2$ and $\ell_{\infty}$). We test two search schema for the trade-off parameter $\lambda$ in optimization: both use $9$ steps of binary search, and $30$ and $1000$ iterations of optimization are performed for each trial of $\lambda$. Only $\alpha=1$ is used in Eqn.~\ref{eq:logit-others} in experiments for simplicity due to computational demand. 
We compare the results under three scenarios proposed in the C\&W method~\cite{CWAttack}: \textit{The Best Case} settings test the attack against all incorrect classes, and report the target class(es) that was least difficult to attack.
\textit{The Worst Case} settings test the attack against all incorrect classes, and report the target class(es) that was most difficult to attack.
\textit{The Average Case} settings select the target class(es) uniformly at random among the labels that are not the GT.

\begin{figure} [h]
  \centering
    \includegraphics[width=1.0\linewidth]{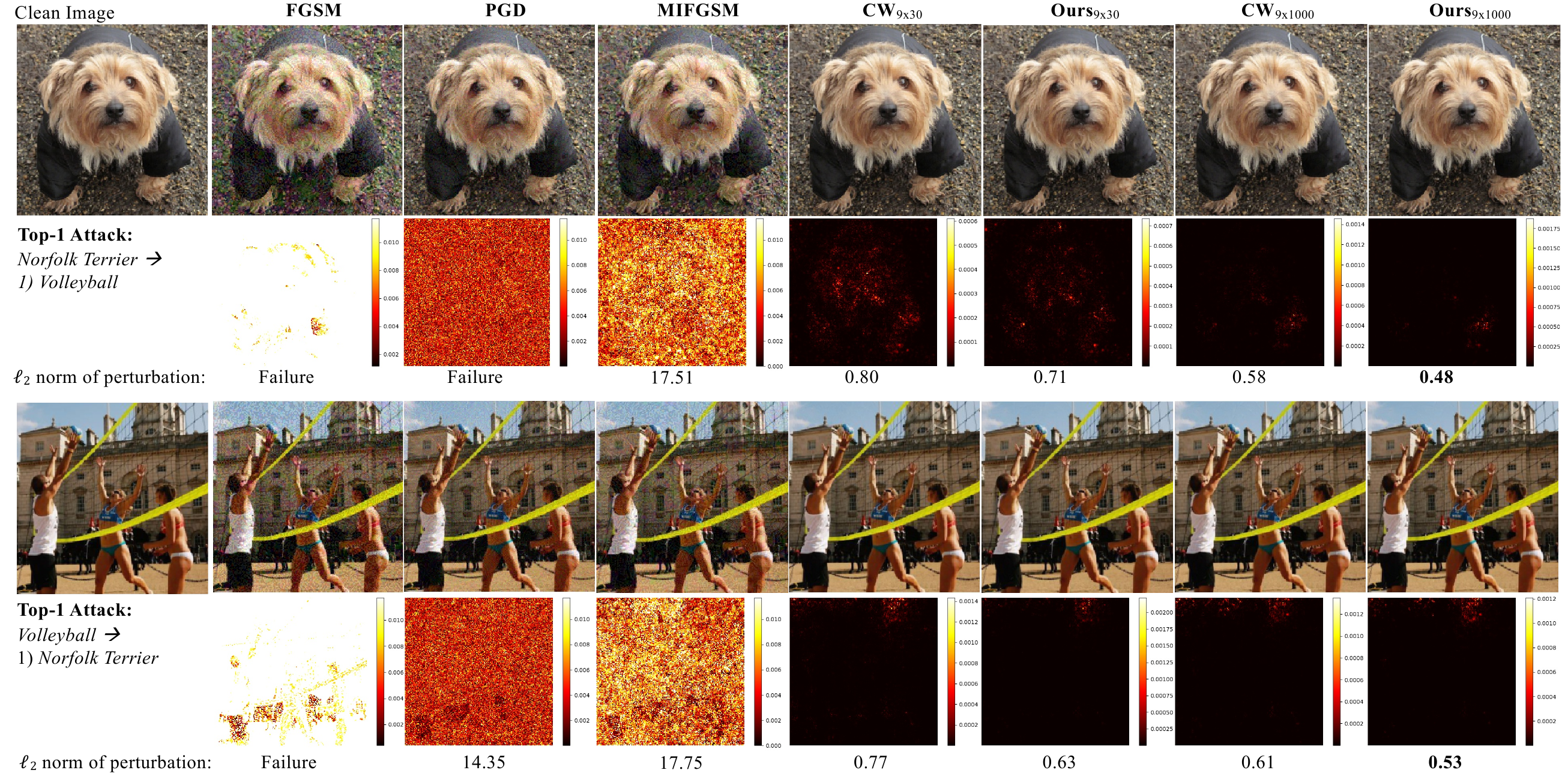}
  \caption{\small Adversarial examples learned with the Top-$1$ attack setting using ResNet-50~\cite{ResidualNet}. The perturbation is shown by $\ell_{2}$ distance between the clean image and adversarial examples. For better visualization, we use different scales in showing the heat maps for different methods. (Best viewed in color and magnification)}\label{fig:ex-top1}
\end{figure}

\begin{table*} [h]
\caption{\small Results and comparisons under the Top-$1$ targeted attack setting. We also test against three state-of-the-art untargeted attack methods, FGSM and PGD and MIFGSM, and the last two use 10 steps in optimization.}
\label{tab:top1}
\centering
\resizebox{1\textwidth}{!}{
\begin{tabular}{llllllllllllll} 
\toprule
\multirow{2}{*}{Model}&\multirow{2}{*}{Attack Method} & \multicolumn{4}{c}{Best Case}&\multicolumn{4}{c}{Average Case} &\multicolumn{4}{c}{Worst Case} \\  
\cmidrule(r){3-14}
&&ASR&$\ell_{1}$&$\ell_{2}$&$\ell_{\infty}$&ASR&$\ell_{1}$&$\ell_{2}$&$\ell_{\infty}$&ASR&$\ell_{1}$&$\ell_{2}$&$\ell_{\infty}$\\
\midrule
\multirow{7}{*}{ResNet-50~\cite{ResidualNet}}&FGSM~\cite{FGSM} &2.3   &9299 &24.1 &0.063 &0.46   &9299 &24.1 &0.063 &0   &N.A. &N.A. &N.A.\\
&PGD$_{10}$~\cite{IFGSM, PGD} &99.6   &4691 &14.1 &0.063 &88.1   &4714 &14.2 &0.063 &57.1   &4748 &14.3 &0.063\\
&MIFGSM$_{10}$~\cite{MIFGSM} &100   &5961 &17.4 &0.063 &99.98   &6082 &17.6 &0.063 &99.9   &6211 &17.9 &0.063\\
&C\&W$_{9\times30}$~\cite{CWAttack} &100   &209.7 &0.777 &0.022 &99.92   &354.1 &1.273 &0.031 &99.9   &560.9 &1.987 &0.042\\
&Ours$_{9\times30}$ &100   &140.9 &0.542 &0.018 &99.9   &184.6 &0.696 &0.025 &99.9   &238.6 &0.880 &0.032\\
&C\&W$_{9\times1000}$~\cite{CWAttack} &100   &95.6 &0.408 &0.017 &100   &127.2 &0.516 &\textbf{0.023} &100   &164.1 &0.635 &0.030\\
&Ours$_{9\times1000}$ &100   &\textbf{81.3} &\textbf{0.380} &\textbf{0.016} &100   &\textbf{109.6} &\textbf{0.472} &\textbf{0.023} &100   &\textbf{143.9} &\textbf{0.579} &\textbf{0.029}\\
\midrule
\multirow{7}{*}{DenseNet-121~\cite{DenseNet}}&FGSM~\cite{FGSM} &6.4   &9263 &24.0 &0.063 &1.44   &9270 &24.0 &0.063 &0   &N.A. &N.A. &N.A.\\
&PGD$_{10}$~\cite{IFGSM, PGD} &100   &4617 &14.2 &0.063 &97.2   &4716 &14.2 &0.063 &87.6   &4716 &14.2 &0.063\\
&MIFGSM$_{10}$~\cite{MIFGSM} &100   &5979 &17.6 &0.063 &100   &6095 &17.6 &0.063 &100   &6218 &17.9 &0.063\\
&C\&W$_{9\times30}$~\cite{CWAttack} &99.9   &188.6 &0.694 &0.019 &99.9   &279.4 &1.008 &0.028 &99.9   &396.5 &1.404 &0.037\\
&Ours$_{9\times30}$ &99.9   &136.4&0.523 &0.017 &99.9   &181.8 &0.678 &0.024 &99.9   &240.0 &0.870 &0.031\\
&C\&W$_{9\times1000}$~\cite{CWAttack} &100   &98.5 &0.415 &\textbf{0.016} &100   &132.3 &0.528 &\textbf{0.023} &100   &174.8 &0.657 &\textbf{0.030}\\
&Ours$_{9\times1000}$ &100   &\textbf{83.8} &\textbf{0.384} &\textbf{0.016} &100   &\textbf{115.9} &\textbf{0.485} &\textbf{0.023} &100   &\textbf{158.69} &\textbf{0.610} &\textbf{0.030}\\
\bottomrule
\end{tabular}
}
\end{table*}

\begin{table*}
\caption{\small Results of Top-$1$ targeted attacks using 5 most-like labels and 5 least-like labels as targets respectively, based on the label semantic similarities.}
\label{tab:top1-sim}
\centering
\resizebox{1\textwidth}{!}{
\begin{tabular}{lllllllllllllllll} 
\toprule
\multirow{2}{*}{Model}&\multirow{2}{*}{Similarity} &\multirow{2}{*}{Attack Method}& \multicolumn{4}{c}{Best Case}&\multicolumn{4}{c}{Average Case} &\multicolumn{4}{c}{Worst Case} \\
\cmidrule(r){4-15}
&&&ASR&$\ell_{1}$&$\ell_{2}$&$\ell_{\infty}$&ASR&$\ell_{1}$&$\ell_{2}$&$\ell_{\infty}$&ASR&$\ell_{1}$&$\ell_{2}$&$\ell_{\infty}$ \ &\\
\midrule
\multirow{14}{*}{ResNet-50~\cite{ResidualNet}}&\multirow{7}{*}{Most like}
&FGSM~\cite{FGSM} &32.4   &9137 &23.8 &0.063 &0.0862   &9137 &23.8 &0.063 &0   &N.A. &N.A. &N.A.\\
&&PGD$_{10}$~\cite{IFGSM, PGD} &99.9   &4687 &14.1 &0.063 &94.6   &4708 &14.2 &0.063 &78.4   &4737 &14.3 &0.063\\
&&MIFGSM$_{10}$~\cite{MIFGSM}&100   &5993 &17.4 &0.063 &99.98   &6110 &17.7 &0.063 &99.9   &6228 &17.9 &0.063 \\
&&C\&W$_{9\times30}$~\cite{CWAttack} &100   &138 &0.51 &0.012 &99.94   &249 &0.89 &0.023 &99.9   &401 &1.40 &0.035\\
&&Ours$_{9\times30}$ &100   &114 &0.43 &0.011 &99.96   &171 &0.62 &0.020 &99.9   &251 &0.87 &0.028\\
&&C\&W$_{9\times1000}$~\cite{CWAttack} &100   &82 &0.34 &\textbf{0.010} &100   &126 &0.48 &\textbf{0.018} &100   &194 &0.68 &0.026\\
&&Ours$_{9\times1000}$&100   &\textbf{75} &\textbf{0.32} &\textbf{0.010} &100   &\textbf{117} &\textbf{0.46} &\textbf{0.018} &100   &\textbf{187} &\textbf{0.65} &\textbf{0.025}\\
\cmidrule(r){2-15}
&\multirow{7}{*}{Least like}
&FGSM~\cite{FGSM} &0.4   &8860 &23.4 &0.063 &0.08   &8860 &23.4 &0.063 &0   &N.A. &N.A. &N.A.\\
&&PGD$_{10}$~\cite{IFGSM, PGD} &99.4   &4696 &14.1 &0.063 &84.82   &4721 &14.2 &0.063 &50   &4762 &14.3 &0.063\\
&&MIFGSM$_{10}$~\cite{MIFGSM}&100   &5960 &17.4 &0.0625 &99.96   &6069 &17.6 &0.063 &99.8   &6194 &17.9 &0.063\\
&&C\&W$_{9\times30}$~\cite{CWAttack}  &99.9   &259 &0.95 &0.025 &99.9   &421 & 1.51&0.035 &99.9   &639 &2.25 &0.046\\
&&Ours$_{9\times30}$ &99.9   &154 &0.59 &0.020 &99.9   &194 & 0.73 &0.026 &99.9   &240 &0.89 &0.033\\
&&C\&W$_{9\times1000}$~\cite{CWAttack} &100   &102 &0.44 &\textbf{0.019} &100   &132 &0.54 &0.025 &100   &165 &0.65 &0.032\\
&&Ours$_{9\times1000}$&100   &\textbf{85} &\textbf{0.40} &\textbf{0.019} &100   &\textbf{111} &\textbf{0.49} &\textbf{0.024} &100   &\textbf{142} &\textbf{0.59} &\textbf{0.030}\\
\midrule
\multirow{14}{*}{DenseNet-121~\cite{DenseNet}}&\multirow{7}{*}{Most like}
&FGSM~\cite{FGSM} &46.1   &9132 &23.8 &0.063 &14.62   &9143 &23.8 &0.063 &0.3  &9263 &24.0. &0.063\\
&&PGD$_{10}$~\cite{IFGSM, PGD} &100   &4692 &14.1 &0.063 &98.92   &4712 &14.2 &0.063 &94.7   &4733 &14.3 &0.063\\
&&MIFGSM$_{10}$~\cite{MIFGSM}&100   &6010 &17.5 &0.063 &100   &6128 &17.7 &0.063 &100   &6245 &18.0 &0.063 \\
&&C\&W$_{9\times30}$~\cite{CWAttack} &100   &130 &0.48 &0.010 &100   &218 &0.77 &0.021 &100   &332 &1.14 &0.031\\
&&Ours$_{9\times30}$ &100   &114 &0.42 &0.010 &100   &170 &0.61 &0.019 &100   &250 &0.85 &0.028\\
&&C\&W$_{9\times1000}$~\cite{CWAttack}&100   &85 &0.34 &\textbf{0.010} &100   &134 &0.50 &\textbf{0.018} &100   &210 &0.71 &\textbf{0.026}\\
&&Ours$_{9\times1000}$&100   &\textbf{77} &\textbf{0.33} &\textbf{0.010} &100   &\textbf{124} &\textbf{0.48} &\textbf{0.018} &100   &\textbf{202} &\textbf{0.69} &\textbf{0.026}\\
\cmidrule(r){2-15}
&\multirow{7}{*}{Least like}
&FGSM~\cite{FGSM} &2.1   &9101 &23.7 &0.063 &0.42   &9101 &23.7 &0.063 &0  &N.A. &N.A. &N.A.\\
&&PGD$_{10}$~\cite{IFGSM, PGD} &100   &4698 &14.2 &0.063 &96.32   &4718 &14.2 &0.063 &83.6   &4745 &14.3 &0.063\\
&&MIFGSM$_{10}$~\cite{MIFGSM}&100   &5973 &17.4 &0.0625 &99.98   &6082 &17.9 &0.063 &99.9   &6203 &17.9 &0.063\\
&&C\&W$_{9\times30}$~\cite{CWAttack}  &99.9   &215 &0.79 &0.023 &99.9   &310 & 1.12&0.030 &99.9   &428 &1.52 &0.039\\
&&Ours$_{9\times30}$ &99.9   &145 &0.56 &0.019 &99.9   &188 & 0.70 &0.025 &99.9   &240 &0.88 &0.032\\
&&C\&W$_{9\times1000}$~\cite{CWAttack} &100   &102 &0.43 &\textbf{0.018} &100   &134 &0.54 &\textbf{0.024} &100   &170 &0.65 &\textbf{0.031}\\
&&Ours$_{9\times1000}$&100   &\textbf{85} &\textbf{0.40} &\textbf{0.018} &100   &\textbf{114} &\textbf{0.49} &\textbf{0.024} &100   &\textbf{149} &\textbf{0.59} &\textbf{0.031}\\

\bottomrule
\end{tabular}
}
\end{table*}

\subsection{Results for the Top-$1$ attack setting}
We first evaluate whether the proposed adversarial distillation framework is effective for the traditional Top-$1$ attack setting. The results show that the proposed method can consistently outperform the C\&W method, as well as some other state-of-the-art untargeted attack methods including the PGD method~\cite{PGD,IFGSM}.

Figure~\ref{fig:ex-top1} shows two qualitative results. We can see the C\&W method and our proposed method ``attend" to different regions in images to achieve the attacks.   Table~\ref{tab:top1} shows the quantitative results. Our proposed method obtains smaller $\ell_1$ and $\ell_2$ norm,  while the $\ell_{\infty}$ norm are almost the same. Note that we only use the $\ell_2$ norm in the objective function in learning. We will evaluate the results of explicitly using $\ell_1$ and $\ell_{\infty}$ norm as penalty respectively in future work.

As shown in Table~\ref{tab:top1-sim}, we also test whether the label semantic knowledge can help identify the weak spots of different attack methods, and whether the proposed method can gain more in those weak spots. We observe that attacks are more challenging if the Top-$1$ target is selected from the least-like set in terms of the label semantic similarity (see Eqn.~\ref{eq:logit-others}). 

\subsection{Results for the ordered Top-$5$ attack setting}
We test ordered Top-$5$ attacks and compare with the modified C\&W method. Our proposed method significantly outperforms the modified C\&W method, especially for the $9\times 30$ optimization scheme, as shown in Table~\ref{tab:top5}. We also observe improvement on the $\ell_{\infty}$ norm for the ordered Top-$5$ attacks (please see Figure~\ref{fig:ex-top5-attack} for two visual examples). 

We also test the effectiveness of knowledge-oriented specifications of selecting the ordered Top-$5$ targets with similar observation obtained as in the Top-$1$ experiments (see Table~\ref{tab:top5-sim}).    

To further evaluate the proposed method, we also test the ordered Top-$5$ attacks using labels with 5 highest and 5 lowest clean prediction scores as targets respectively, as shown in Table~\ref{tab:top5-clean-logit}. We observe similar patterns that the 5 labels with the lowest clean prediction scores are more challenging to attack. This shed lights on learning data-driven knowledge: Instead of using the label semantic knowledge which may have some discrepancy in guiding the design of adversarial loss functions, we can leverage the similarities measured based on the confusion matrix in the training data if available. We leave this for future work. 

\begin{table*}
\caption{\small Results and comparisons under the ordered Top-$5$ targeted attack protocol using randomly selected and ordered 5 targets (GT exclusive).}
\label{tab:top5}
\centering
\resizebox{1\textwidth}{!}{
\begin{tabular}{llllllllllllll} 
\toprule
\multirow{2}{*}{Model}&\multirow{2}{*}{Attack Method} & \multicolumn{4}{c}{Best Case}&\multicolumn{4}{c}{Average Case} &\multicolumn{4}{c}{Worst Case} \\  
\cmidrule(r){3-14}
&&ASR&$\ell_{1}$&$\ell_{2}$&$\ell_{\infty}$&ASR&$\ell_{1}$&$\ell_{2}$&$\ell_{\infty}$&ASR&$\ell_{1}$&$\ell_{2}$&$\ell_{\infty}$\\
\midrule
\multirow{4}{*}{ResNet-50~\cite{ResidualNet}}&C\&W$_{9\times30}$~\cite{CWAttack} &75.8   &2370 &7.76 &0.083 &29.34   &2425 &7.94 &0.086 &0.7   &2553 &8.37 &0.094\\
&Ours$_{9\times30}$ &96.1   &1060 &3.58 &0.056 &80.68   &1568 &5.13 &0.070 &49.8   &2215 &7.07 &0.087\\
&C\&W$_{9\times1000}$~\cite{CWAttack} &100   &437 &1.59 &0.044 &100   &600 &2.16 &0.058 &100   &779 &2.77 &0.074\\
&Ours$_{9\times1000}$ &100   &\textbf{285} &\textbf{1.09} &\textbf{0.034} &100   &\textbf{359} &\textbf{1.35} &\textbf{0.043} &100   &\textbf{456} &\textbf{1.68} &\textbf{0.055}\\
\midrule
\multirow{4}{*}{DenseNet-121~\cite{DenseNet}}&C\&W$_{9\times30}$~\cite{CWAttack} &96.6   &2161 &7.09 &0.071 &73.68   &2329 &7.65 &0.080 &35.6   &2530 &8.28 &0.088\\
&Ours$_{9\times30}$ &97.7   &6413 &2.14 &0.043 &92.66   &1063 &3.57 &0.057 &83.3   &1636 &5.35 &0.072\\
&C\&W$_{9\times1000}$~\cite{CWAttack} &100  &392 &1.42 &0.040 &100   &527 &1.89 &0.052&100   &669 &2.37 &0.065\\
&Ours$_{9\times1000}$ &100   &\textbf{273} &\textbf{1.05} &\textbf{0.033} &100   &\textbf{344} &\textbf{1.29} &\textbf{0.042} &100   &\textbf{425} &\textbf{1.57} &\textbf{0.052}\\
\bottomrule
\end{tabular}
}
\end{table*}

\begin{table}
\caption{\small Results of ordered Top-$5$ targeted attacks using 5 most-like labels and 5 least-like labels as targets respectively, based on the label semantic similarities.}
\label{tab:top5-sim}
\centering
\resizebox{0.8\textwidth}{!}{
\begin{tabular}{lllllll} 
\toprule
Model&Similarity &Attack Method&ASR&$\ell_{1}$&$\ell_{2}$&$\ell_{\infty}$\\
\midrule
\multirow{8}{*}{ResNet-50~\cite{ResidualNet}}&\multirow{4}{*}{Most like}
&C\&W$_{9\times30}$~\cite{CWAttack} &80   &1922 &6.30 &0.066\\
&&Ours$_{9\times30}$ &96.5   &1286 &4.20 &0.054 \\
&&C\&W$_{9\times1000}$~\cite{CWAttack}&100   &392 &1.43 &0.042 \\
&&Ours$_{9\times1000}$&100   &\textbf{277} &\textbf{1.05} &\textbf{0.035} \\
\cmidrule(r){2-7}
&\multirow{4}{*}{Least like}
&C\&W$_{9\times30}$~\cite{CWAttack} &27.1   &2418 &7.90 &0.085 \\
&&Ours$_{9\times30}$ &77.1   &1635 &5.35 &0.072\\
&&C\&W$_{9\times1000}$~\cite{CWAttack} &100   &596 &2.15 &0.060 \\
&&Ours$_{9\times1000}$ &100   &\textbf{370} &\textbf{1.39} &\textbf{0.045} \\
\midrule
\multirow{8}{*}{DenseNet-121~\cite{DenseNet}}&\multirow{4}{*}{Most like}
&C\&W$_{9\times30}$~\cite{CWAttack} &92.1   &1798 &5.88 &0.059\\
&&Ours$_{9\times30}$ &98.4   &1228 &4.00 &0.050 \\
&&C\&W$_{9\times1000}$~\cite{CWAttack}&100   &361 &1.31 &0.039 \\
&&Ours$_{9\times1000}$&100   &\textbf{265} &\textbf{1.00} &\textbf{0.034} \\
\cmidrule(r){2-7}
&\multirow{4}{*}{Least like}
&C\&W$_{9\times30}$~\cite{CWAttack} &75.7   &2325 &7.64 &0.080 \\
&&Ours$_{9\times30}$ &92.8   &1076 &3.63 &0.057\\
&&C\&W$_{9\times1000}$~\cite{CWAttack} &100   &529 &1.90 &0.052 \\
&&Ours$_{9\times1000}$&100   &\textbf{343} &\textbf{1.29} &\textbf{0.042} \\
\bottomrule
\end{tabular}
}
\end{table}



\begin{table}
\caption{\small Results of ordered Top-$5$ targeted attacks using labels with 5 highest and 5 lowest  prediction scores of clean images as targets respectively. }
\label{tab:top5-clean-logit}
\centering
\resizebox{0.8\textwidth}{!}{
\begin{tabular}{lllllll} 
\toprule
Model&Clean prediction &Attack Method&ASR&$\ell_{1}$&$\ell_{2}$&$\ell_{\infty}$\\
\midrule
\multirow{8}{*}{ResNet-50~\cite{ResidualNet}}&\multirow{4}{*}{Highest}
&C\&W$_{9\times30}$~\cite{CWAttack} &93   &1546 &4.98 &0.042\\
&&Ours$_{9\times30}$ &99.9   &1182 &3.78 &0.039 \\
&&C\&W$_{9\times1000}$~\cite{CWAttack}&100   &205 &0.75 &0.025 \\
&&Ours$_{9\times1000}$&100   &\textbf{170} &\textbf{0.65} &\textbf{0.023} \\
\cmidrule(r){2-7}
&\multirow{4}{*}{Lowest}
&C\&W$_{9\times30}$~\cite{CWAttack} &13.4  &2231 &7.30 &0.082 \\
&&Ours$_{9\times30}$ &68.6   &1791 &5.86&0.077\\
&&C\&W$_{9\times1000}$~\cite{CWAttack} &100   &621 &2.25 &0.064 \\
&&Ours$_{9\times1000}$ &100   &\textbf{392} &\textbf{1.47} &\textbf{0.047} \\
\bottomrule
\end{tabular}
}
\end{table}

\section{Related work}\label{sec:related}
The growing ubiquity of DNNs in advanced machine learning and AI systems dramatically increases their capabilities, but also increases the potential for new vulnerabilities to attacks. This situation has become critical as many powerful approaches have been developed where imperceptible perturbations to DNN inputs could deceive a well-trained DNN, significantly altering its prediction. 
Please refer to~\cite{AttackSurvey} for a comprehensive survey of attack methods in computer vision. We review some related work that motivate our work and show the difference. 

\textbf{Distillation.}
The central idea of our proposed work is built on distillation. Network distillation~\cite{distillation1,distillation} is a powerful training scheme proposed to train a new, usually lightweight model (a.k.a., the student) to mimic another already trained model (a.k.a. the teacher). It takes a functional viewpoint of the knowledge learned by the teacher as the conditional distribution it produces over outputs given an input. It teaches the student to keep up or emulate by adding some regularization terms to the loss in order to encourage the two models to be similar directly based on the distilled knowledge, replacing the training labels. Label smoothing~\cite{LabelSmoothing} can be treated as a simple hand-crafted knowledge to help improve model performance. 
Distillation has been exploited to develop defense models~\cite{distillation_defense} to improve model robustness. Our proposed adversarial distillation method utilizes the distillation idea in an opposite direction, leveraging label semantic driven knowledge for learning ordered Top-$k$ attacks and improving attack robustness.  

\textbf{Adversarial Attack.} For image classification tasks using DNNs, the discovery of the existence of visually-imperceptible adversarial attacks~\cite{LBFGS} was a big shock in developing DNNs. 
White-box attacks provide a powerful way of evaluating model brittleness. In a plain and loose explanation, DNNs are universal function approximator~\cite{UniversalApproximator} and capable of even fitting random labels~\cite{DNNGeneralization} in large scale classification tasks as ImageNet-1000~\cite{ImageNet}. Thus, adversarial attacks are always learnable provided proper objective functions are given, especially when DNNs are trained with fully differentible back-propagation. Many white-box attack methods focus on norm-ball constrained objective functions~\cite{LBFGS,IFGSM,CWAttack,MIFGSM}. The C\&W method investigates 7 different loss functions. The best performing loss function found by the C\&W method has been appliedin many attack methods and achieved strong results~\cite{ZOOAttack, PGD, EADAttack}. By introducing momentum in the MIFGSM method~\cite{MIFGSM} and the $\ell_{p}$ gradient projection in the PGD method~\cite{PGD}, they usually achieve better performance in generating adversarial examples. In the meanwhile, some other attack methods such as the StrAttack~\cite{StrAttack} also investigate different loss functions for better interpretability of attacks. Our proposed method leverage label semantic knowledge in the loss function design for the first time.


\section{Conclusions}\label{sec:conclusion}
This paper proposes to extend the traditional Top-$1$ targeted attack setting to the ordered Top-$k$ setting ($k\geq 1$) under the white-box attack protocol. The ordered Top-$k$ targeted attacks can improve the robustness of attacks themselves. To our knowledge, it is the first work studying this ordered Top-$k$ attacks. To learn the ordered Top-$k$ attacks, we present a conceptually simple yet effective adversarial distillation framework motivated by network distillation. We also develop a modified C\&W method as the strong baseline for the ordered Top-$k$ targeted attacks. In experiments, the proposed method is tested in ImageNet-1000 using two popular DNNs, ResNet-50 and DenseNet-121, with consistently better results obtained. We investigate the effectiveness of label semantic knowledge in designing the adversarial distribution for distilling the ordered Top-$k$ targeted attacks.

\section*{Acknowledgments} 
This work is supported by ARO grant W911NF1810295 and ARO DURIP grant W911NF1810209, and NSF IIS 1822477.



\small
\bibliography{reference}

\begin{thebibliography}{28}
\providecommand{\natexlab}[1]{#1}
\providecommand{\url}[1]{\texttt{#1}}
\expandafter\ifx\csname urlstyle\endcsname\relax
  \providecommand{\doi}[1]{doi: #1}\else
  \providecommand{\doi}{doi: \begingroup \urlstyle{rm}\Url}\fi

\bibitem[Carlini and Wagner(2016)]{CWAttack}
Nicholas Carlini and David~A. Wagner.
\newblock Towards evaluating the robustness of neural networks.
\newblock \emph{CoRR}, abs/1608.04644, 2016.
\newblock URL \url{http://arxiv.org/abs/1608.04644}.

\bibitem[Hinton et~al.(2015)Hinton, Vinyals, and Dean]{distillation}
Geoffrey~E. Hinton, Oriol Vinyals, and Jeffrey Dean.
\newblock Distilling the knowledge in a neural network.
\newblock \emph{CoRR}, abs/1503.02531, 2015.
\newblock URL \url{http://arxiv.org/abs/1503.02531}.

\bibitem[Russakovsky et~al.(2015)Russakovsky, Deng, Su, Krause, Satheesh, Ma,
  Huang, Karpathy, Khosla, Bernstein, Berg, and Fei-Fei]{ImageNet}
Olga Russakovsky, Jia Deng, Hao Su, Jonathan Krause, Sanjeev Satheesh, Sean Ma,
  Zhiheng Huang, Andrej Karpathy, Aditya Khosla, Michael Bernstein,
  Alexander~C. Berg, and Li~Fei-Fei.
\newblock {ImageNet Large Scale Visual Recognition Challenge}.
\newblock \emph{International Journal of Computer Vision (IJCV)}, 115\penalty0
  (3):\penalty0 211--252, 2015.
\newblock \doi{10.1007/s11263-015-0816-y}.

\bibitem[He et~al.(2016)He, Zhang, Ren, and Sun]{ResidualNet}
Kaiming He, Xiangyu Zhang, Shaoqing Ren, and Jian Sun.
\newblock Deep residual learning for image recognition.
\newblock In \emph{IEEE Conference on Computer Vision and Pattern Recognition
  (CVPR)}, 2016.

\bibitem[Huang et~al.(2017)Huang, Liu, van~der Maaten, and
  Weinberger]{DenseNet}
Gao Huang, Zhuang Liu, Laurens van~der Maaten, and Kilian~Q Weinberger.
\newblock Densely connected convolutional networks.
\newblock In \emph{Proceedings of the IEEE Conference on Computer Vision and
  Pattern Recognition}, 2017.

\bibitem[LeCun et~al.(1998)LeCun, Bottou, Bengio, and Haffner]{LeCunCNN}
Yann LeCun, Leon Bottou, Yoshua Bengio, and Patrick Haffner.
\newblock Gradient-based learning applied to document recognition.
\newblock \emph{Proceedings of the IEEE}, 86\penalty0 (11):\penalty0
  2278--2324, 1998.

\bibitem[Krizhevsky et~al.(2012)Krizhevsky, Sutskever, and Hinton]{AlexNet}
Alex Krizhevsky, Ilya Sutskever, and Geoffrey~E. Hinton.
\newblock Imagenet classification with deep convolutional neural networks.
\newblock In \emph{Neural Information Processing Systems (NIPS)}, pages
  1106--1114, 2012.

\bibitem[Szegedy et~al.(2016)Szegedy, Ioffe, and Vanhoucke]{InceptionNet}
Christian Szegedy, Sergey Ioffe, and Vincent Vanhoucke.
\newblock Inception-v4, inception-resnet and the impact of residual connections
  on learning.
\newblock \emph{CoRR}, abs/1602.07261, 2016.
\newblock URL \url{http://arxiv.org/abs/1602.07261}.

\bibitem[Nguyen et~al.(2015)Nguyen, Yosinski, and Clune]{FoolDeepNet}
Anh~Mai Nguyen, Jason Yosinski, and Jeff Clune.
\newblock Deep neural networks are easily fooled: High confidence predictions
  for unrecognizable images.
\newblock In \emph{{IEEE} Conference on Computer Vision and Pattern
  Recognition, {CVPR} 2015, Boston, MA, USA, June 7-12, 2015}, pages 427--436,
  2015.
\newblock URL \url{http://dx.doi.org/10.1109/CVPR.2015.7298640}.

\bibitem[Szegedy et~al.(2014)Szegedy, Zaremba, Sutskever, Bruna, Erhan,
  Goodfellow, and Fergus]{LBFGS}
Christian Szegedy, Wojciech Zaremba, Ilya Sutskever, Joan Bruna, Dumitru Erhan,
  Ian~J. Goodfellow, and Rob Fergus.
\newblock Intriguing properties of neural networks.
\newblock In \emph{{ICLR}}, 2014.

\bibitem[Athalye and Sutskever(2017)]{AdversarialExampl}
Anish Athalye and Ilya Sutskever.
\newblock Synthesizing robust adversarial examples.
\newblock \emph{CoRR}, abs/1707.07397, 2017.
\newblock URL \url{http://arxiv.org/abs/1707.07397}.

\bibitem[Su et~al.(2017)Su, Vargas, and Kouichi]{OnePixelAttack}
Jiawei Su, Danilo~Vasconcellos Vargas, and Sakurai Kouichi.
\newblock One pixel attack for fooling deep neural networks.
\newblock \emph{CoRR}, abs/1710.08864, 2017.

\bibitem[Ilyas et~al.(2019)Ilyas, Santurkar, Tsipras, Engstrom, Tran, and
  Madry]{AttackNotBugs}
Andrew Ilyas, Shibani Santurkar, Dimitris Tsipras, Logan Engstrom, Brandon
  Tran, and Aleksander Madry.
\newblock Adversarial examples are not bugs, they are features.
\newblock \emph{CoRR}, abs/1905.02175, 2019.
\newblock URL \url{http://arxiv.org/abs/1905.02175}.

\bibitem[Szegedy et~al.(2015)Szegedy, Vanhoucke, Ioffe, Shlens, and
  Wojna]{LabelSmoothing}
Christian Szegedy, Vincent Vanhoucke, Sergey Ioffe, Jonathon Shlens, and
  Zbigniew Wojna.
\newblock Rethinking the inception architecture for computer vision.
\newblock \emph{CoRR}, abs/1512.00567, 2015.
\newblock URL \url{http://arxiv.org/abs/1512.00567}.

\bibitem[Pereyra et~al.(2017)Pereyra, Tucker, Chorowski, Kaiser, and
  Hinton]{ConfidencePenalty}
Gabriel Pereyra, George Tucker, Jan Chorowski, Lukasz Kaiser, and Geoffrey~E.
  Hinton.
\newblock Regularizing neural networks by penalizing confident output
  distributions.
\newblock \emph{CoRR}, abs/1701.06548, 2017.
\newblock URL \url{http://arxiv.org/abs/1701.06548}.

\bibitem[Bucila et~al.(2006)Bucila, Caruana, and
  Niculescu{-}Mizil]{distillation1}
Cristian Bucila, Rich Caruana, and Alexandru Niculescu{-}Mizil.
\newblock Model compression.
\newblock In \emph{Proceedings of the Twelfth {ACM} {SIGKDD} International
  Conference on Knowledge Discovery and Data Mining, Philadelphia, PA, USA,
  August 20-23, 2006}, pages 535--541, 2006.
\newblock \doi{10.1145/1150402.1150464}.
\newblock URL \url{https://doi.org/10.1145/1150402.1150464}.

\bibitem[Goodfellow et~al.(2015)Goodfellow, Shlens, and Szegedy]{FGSM}
Ian Goodfellow, Jonathon Shlens, and Christian Szegedy.
\newblock Explaining and harnessing adversarial examples.
\newblock In \emph{International Conference on Learning Representations}, 2015.
\newblock URL \url{http://arxiv.org/abs/1412.6572}.

\bibitem[Kurakin et~al.(2017)Kurakin, Goodfellow, and Bengio]{IFGSM}
Alexey Kurakin, Ian~J. Goodfellow, and Samy Bengio.
\newblock Adversarial examples in the physical world.
\newblock In \emph{{ICLR} (Workshop)}. OpenReview.net, 2017.

\bibitem[Madry et~al.(2018)Madry, Makelov, Schmidt, Tsipras, and Vladu]{PGD}
Aleksander Madry, Aleksandar Makelov, Ludwig Schmidt, Dimitris Tsipras, and
  Adrian Vladu.
\newblock Towards deep learning models resistant to adversarial attacks.
\newblock In \emph{{ICLR}}. OpenReview.net, 2018.

\bibitem[Dong et~al.(2018)Dong, Liao, Pang, Su, Zhu, Hu, and Li]{MIFGSM}
Yinpeng Dong, Fangzhou Liao, Tianyu Pang, Hang Su, Jun Zhu, Xiaolin Hu, and
  Jianguo Li.
\newblock Boosting adversarial attacks with momentum.
\newblock In \emph{{CVPR}}, pages 9185--9193. {IEEE} Computer Society, 2018.

\bibitem[Pennington et~al.(2014)Pennington, Socher, and Manning]{Glove}
Jeffrey Pennington, Richard Socher, and Christopher~D. Manning.
\newblock Glove: Global vectors for word representation.
\newblock In \emph{In EMNLP}, 2014.

\bibitem[Akhtar and Mian(2018)]{AttackSurvey}
Naveed Akhtar and Ajmal Mian.
\newblock Threat of adversarial attacks on deep learning in computer vision:
  {A} survey.
\newblock \emph{CoRR}, abs/1801.00553, 2018.
\newblock URL \url{http://arxiv.org/abs/1801.00553}.

\bibitem[Papernot et~al.(2016)Papernot, McDaniel, Wu, Jha, and
  Swami]{distillation_defense}
Nicolas Papernot, Patrick~D. McDaniel, Xi~Wu, Somesh Jha, and Ananthram Swami.
\newblock Distillation as a defense to adversarial perturbations against deep
  neural networks.
\newblock In \emph{{IEEE} Symposium on Security and Privacy}, pages 582--597.
  {IEEE} Computer Society, 2016.

\bibitem[Hornik et~al.(1989)Hornik, Stinchcombe, and
  White]{UniversalApproximator}
Kurt Hornik, Maxwell~B. Stinchcombe, and Halbert White.
\newblock Multilayer feedforward networks are universal approximators.
\newblock \emph{Neural Networks}, 2\penalty0 (5):\penalty0 359--366, 1989.
\newblock \doi{10.1016/0893-6080(89)90020-8}.
\newblock URL \url{https://doi.org/10.1016/0893-6080(89)90020-8}.

\bibitem[Zhang et~al.(2016)Zhang, Bengio, Hardt, Recht, and
  Vinyals]{DNNGeneralization}
Chiyuan Zhang, Samy Bengio, Moritz Hardt, Benjamin Recht, and Oriol Vinyals.
\newblock Understanding deep learning requires rethinking generalization.
\newblock \emph{CoRR}, abs/1611.03530, 2016.
\newblock URL \url{http://arxiv.org/abs/1611.03530}.

\bibitem[Chen et~al.(2017)Chen, Zhang, Sharma, Yi, and Hsieh]{ZOOAttack}
Pin{-}Yu Chen, Huan Zhang, Yash Sharma, Jinfeng Yi, and Cho{-}Jui Hsieh.
\newblock {ZOO:} zeroth order optimization based black-box attacks to deep
  neural networks without training substitute models.
\newblock In \emph{AISec@CCS}, pages 15--26. {ACM}, 2017.

\bibitem[Chen et~al.(2018)Chen, Sharma, Zhang, Yi, and Hsieh]{EADAttack}
Pin{-}Yu Chen, Yash Sharma, Huan Zhang, Jinfeng Yi, and Cho{-}Jui Hsieh.
\newblock {EAD:} elastic-net attacks to deep neural networks via adversarial
  examples.
\newblock In \emph{{AAAI}}, pages 10--17. {AAAI} Press, 2018.

\bibitem[Xu et~al.(2018)Xu, Liu, Zhao, Chen, Zhang, Erdogmus, Wang, and
  Lin]{StrAttack}
Kaidi Xu, Sijia Liu, Pu~Zhao, Pin{-}Yu Chen, Huan Zhang, Deniz Erdogmus, Yanzhi
  Wang, and Xue Lin.
\newblock Structured adversarial attack: Towards general implementation and
  better interpretability.
\newblock \emph{CoRR}, abs/1808.01664, 2018.

\end{thebibliography}

\end{document}